\documentclass[letterpaper, 10 pt, conference]{ieeeconf}  

\IEEEoverridecommandlockouts                              

\usepackage{graphicx} 

\usepackage{mathptmx} 
\usepackage{amsmath} 
\usepackage{amssymb}  
\usepackage{amsfonts}
\usepackage{hyperref}

\usepackage{xcolor} 

\title{\LARGE \bf
A Dataset for Developing and Benchmarking Active Vision}

\author{Phil Ammirato$^{1}$, Patrick Poirson$^{1}$, Eunbyung Park$^{1}$, Jana Ko\v{s}eck\'{a}$^{2}$, Alexander C. Berg$^{1}$
\thanks{*Supported by NSF NRI 1527208,1526367 \& NSF 1452851 \& 1446631 }
\thanks{$^{1}$University of North Carolina at Chapel Hill, Computer Science
{\tt\small [ammirato, poirson, eunbyung, aberg]@cs.unc.edu}}%
\thanks{$^{2}$ George Mason University, Computer Science
        {\tt\small kosecka@gmu.edu}}%
}

\begin{document}

\maketitle
\thispagestyle{empty}
\pagestyle{empty}

\begin{abstract}

We present a new public dataset with a focus on simulating robotic vision tasks
in everyday indoor environments using real imagery.  The dataset
includes 20,000+ RGB-D images and 50,000+ 2D bounding
boxes of object instances densely captured in 9 unique scenes.  
We train a fast object category detector for instance detection on our data. 
Using the dataset we show that, although increasingly
accurate and fast, the state of the art for object detection is still
severely impacted by object scale, occlusion, and viewing direction
all of which matter for robotics applications.  We next validate the dataset
for simulating active vision, and use the dataset to develop and
evaluate a deep-network-based system for next best move prediction for
object classification using reinforcement learning. Our dataset is available for download at  \url{cs.unc.edu/~ammirato/active_vision_dataset_website/}.

\end{abstract}

\section{Introduction}
     

The ability to recognize objects is a core functionality for  robots
operating in everyday human environments.  While there has been
amazing recent progress in computer vision on object classification
and detection, especially with deep models, these lines of work do not
address some of the core needs of vision for robotics.  Partly this is
due to biases in the imagery considered and the fact that these
recognition challenges are performed in isolation for each image.  In
robotic applications, the biases are different and recognition is
performed over multiple images, often with active control of the sensing
platform (active vision).  This paper attempts to address part of this disconnect by
introducing a new approach to studying active vision for robotics by
collecting very dense imagery of scenes in order to allow simulating a
robot moving through an environment by sampling appropriate imagery.

The goals are two-fold, to {\em provide a research and development
resource for computer vision} without requiring access to robots for
experiments, and to {\em provide a way to benchmark and compare
different approaches to active vision} without the difficulty and
expense of evaluating the algorithms on the same physical robotics
testbed.
          \begin{figure}[thpb]
      \centering
     \includegraphics[width=1.0\linewidth]{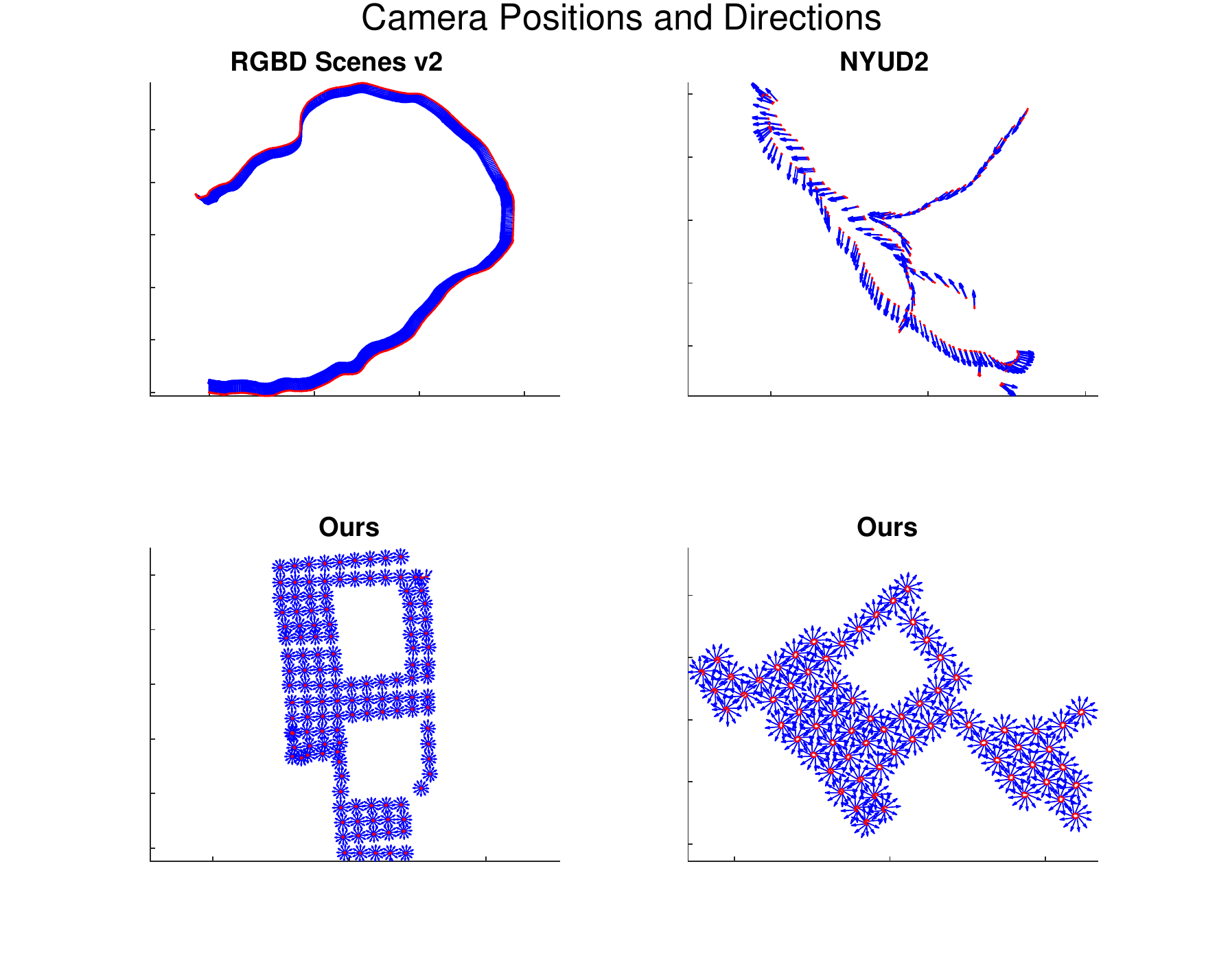}
            \caption{Visualization of camera locations (red) and viewing directions (blue) from our collections (bottom) and previous datasets (top). We collect densely sampled RGB-D images of scenes for use in training and benchmarking active vision systems.  The dense sampling allows ``virtually'' moving a camera through a scene.  While other datasets do sample multiple images per scene, they often sample either from just a few positions or along only a few paths through the environment ~\cite{WASH-RGBDv2,NYUD2}.    Note that the physical scale is different in each plot.}

      \label{fig:cam_poses}
   \end{figure}
We begin by collecting a large dataset of dense RGB-D imagery of common
everyday rooms: kitchens, living rooms, dining rooms, offices, etc.
This imagery is registered and used to form a 3D reconstruction of each
scene.  This reconstruction is used to simplify labeling of objects
in the collection in 3D as opposed to individually in the thousands of
images of those objects.  The geometric relationship between images is
also used to define connectivity for determining what image would be
seen next when moving in a given direction from a given camera position
(e.g. what would I see if I turned right? went backwards?).

Given this labeled data we adapt a state-of-the-art fast object category detector~\cite{liu15ssd} based on deep convolutional networks to the task of recognizing specific object instances in the dataset. While most deep-learning approaches have focused on category detection, instance detection can be practically useful for robotics.  This distinction between recognizing a category of object, such as chair, versus a specific object, such as a particular 8.4oz Red Bull can is important. 
  Our results show that the category detection framework can be adapted to instance detection well, with some caveats.

Where the detection framework has difficulty is in the range of scales, viewing directions, and occlusions present in everyday scenes (e.g. our data) that is different from the biases present in Internet collected datasets.  While the detector performs well for large frontal views of objects its performance falls for other views.  This is quantified in Sec.~\ref{sec:detection}. This view-dependent variation in recognition performance motivates active-vision for object recognition,
controlling the sensing platform to acquire imagery that improves recognition accuracy.

Our high-level goals are based on using the pre-collected dense
imagery to develop and test active-vision algorithms.  To validate
this approach we begin by demonstrating that the imagery is sampled densely enough. 
 In particular
we care that the results and accuracy of recognition algorithms on
samples of the densely collected imagery are close to the results that
would be achieved if the robot moved continuously through the environment. This is explored in Sec.~\ref{sec:density}.

Given this validation, we proceed to use the densely sampled dataset
to train and evaluate a deep-network for determining the next best
move to improve object classification. The recognition component for this is pre-trained with external
data and then a combined network that performs recognition and selects
a direction to move in to improve accuracy is trained on a subset of
the densely sampled data using reinforcement learning.  To illustrate
one way to use the dataset, we employ multiple train/test splits to
determine the expected increase in accuracy with multiple moves using our
next best move network.  See Sec.~\ref{sec:nbv}.

The collected dataset and labels are available at  \url{http://cs.unc.edu/~ammirato/active_vision_dataset_website/}, as well as a small toolbox for visualizations and loading. 
We hope to also provide the functionality to allow groups to submit
algorithms for evaluation on completely private test data in the
future.  Before collection of imagery, release forms were signed and
collected allowing free and legal access to the collected data.

\section{Related Work}  
   This paper proposes an approach to collecting and using datasets to train and benchmark object detection and recognition, especially for {\em active recognition}.   We briefly discuss some of the most related work in each area.


The datasets that have been a driving force in pushing the deep
 learning revolution in object recognition, Pascal VOC~\cite{everingham2010pascal},
 the ImageNet Challenge~\cite{russakovsky2015imagenet}, and MS COCO~\cite{lin2014microsoft}
 are all collected from web images (usually from Flickr) using web
 search based on keywords.  These image collections introduce biases from the
 human photographer, the human tagging, and the web search engine. As
 a result objects are usually of medium to large size in images and
 are usually frontal views with small amounts of occlusion.  In addition
 these datasets focus on object category recognition.  The state of
 the art for object classification and recognition in these datasets
 is based on either object proposals and feature pooling
 following~\cite{uijlings2013selective} with advanced deep
 networks~\cite{girshick2014rich,He_2016} or on fully convolutional networks
 implementing a modern take on sliding windows~\cite{liu15ssd,redmon2015you,ssd-pose} that provide frame-rate or faster performance on high-end hardware for
 some reduction in accuracy.

      Instance recognition (as opposed to object category recognition) has generally been approached using local features or template matching techniques.  A recent relevant example using these types of models is \cite{RobotInARoom} that trains on objects in a room and is tested on the same objects in the room after rearrangement.   In our experiments we are interested in generalization to new environments in order to avoid training in each new room.  More recently, \cite{instanceHeld} shows how deep-learning for comparing instances can be applied to instance classification and outperform classic matching methods. For our data, we are also interested in instance detection, including localization in a large image. We use the system from \cite{liu15ssd} to build a much faster detector for object instances than would be possible with explicit matching.
      



There are many RGB-D datasets available today, but none with a focus on simulating robot motion through an environment. \cite{all-rgbd_datasets} gives a list of a various RGB-D datasets, some  focus on single objects \cite{BigBIRD,UWASH-OBJECTS}, in what we call ``table-top'' style data. 
	This type of data, especially the data in BigBIRD \cite{BigBIRD}, is similar to what manufactures may provide for robots in the future. While not capturing real-world scenes, the number of views and detail for each instance in this data can provide valuable training data for instance recognition systems. We include over 30 object instances similar to those in the BigBIRD dataset in our scenes. 

    

    Scene dataset,  \cite{NYUD2}, \cite{SUN-RGBD},\cite{WASH-RGBDv2}, and \cite{gmu-multiview} do explore environments more than ``table-top'' data but do not have a dense set of views to simulate robot motion. 
    These data-sets often have only one or two paths through the scene. An actual robot in the real-world has many choices of where to move, and the controller has to be able to pick a good path. See Figure~\ref{fig:cam_poses} for a comparison of the available paths through scenes in previous datasets and our data.

    Active vision has a long history in robotics. Early work largely centered around view selection~\cite{bajcsy1988active, jia2011robotic}. Others ~\cite{karasev2012controlled, atanasov2013hypothesis, velez2012modelling} have worked on the problem from a more theoretical perspective, but under many simplified settings for possible motions, or assumptions about known object models. In recent years, next best view prediction has been one of the more popular active vision problems. However, most of these approaches use CAD models of the objects of interest \cite{3dShapeNets,su15mvcnn,Pairwise_active}, with some  small sets of real-world images \cite{NBV-6DOF}.   CAD models produce encouraging results, but leave out some real-world challenges in perception.
    
    
    \cite{NBV-6DOF} gives a system for object detection, pose estimation, and next best view prediction. They are able to test their detection and pose estimation system on existing real image datasets, but need to collect their own data to test their active vision framework. They collect a small scale dataset of only ``table-top'' style scenes with about 30-60 images each. This shows the need for a dataset for active vision, while also showing how difficult it can be to collect such data at a large scale. 

       	          \begin{figure}[thpb]
      \centering
     \includegraphics[width=1.0\linewidth]{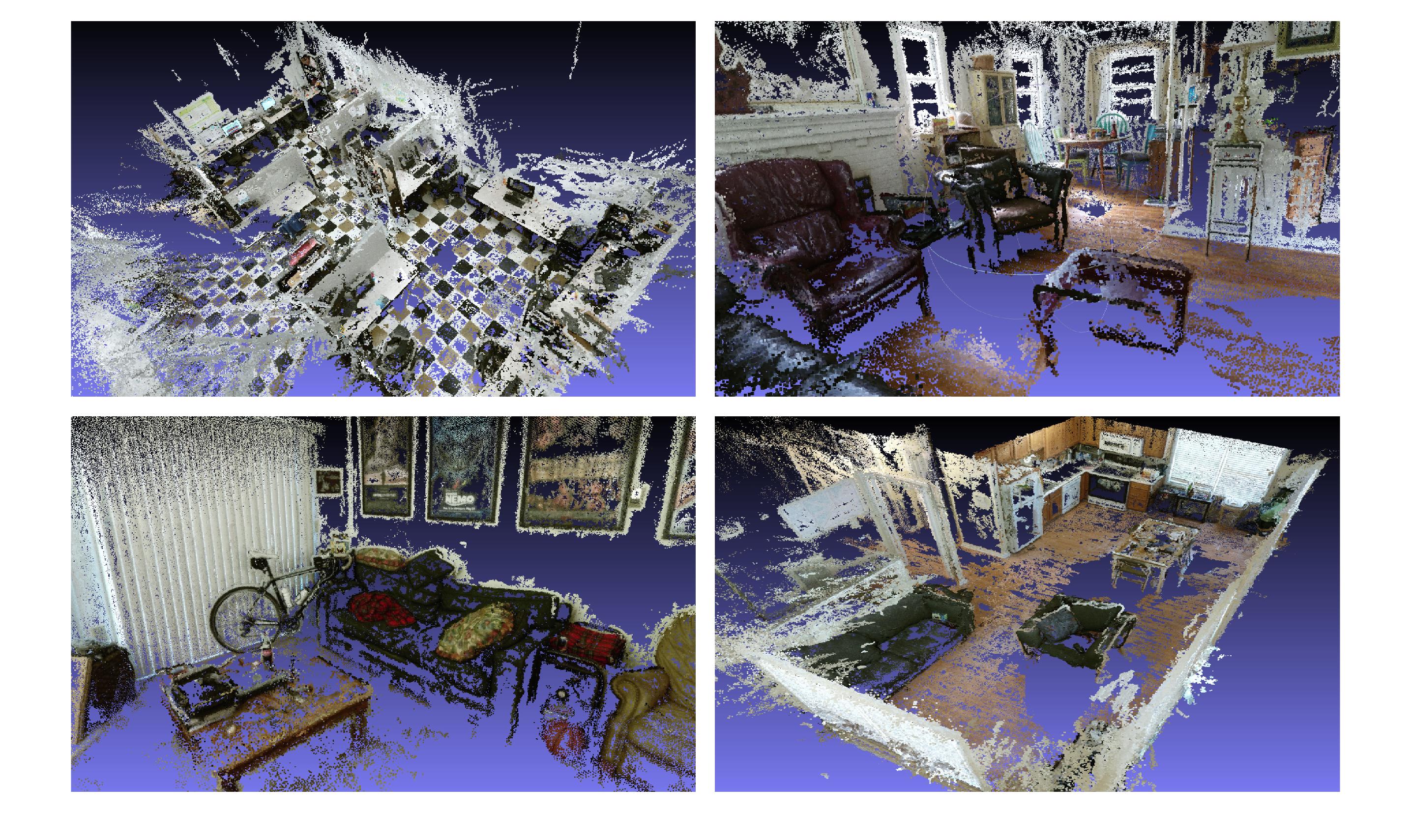}
      \caption{Four dense reconstructions of scenes from our collected data.  We label objects in 3D using the dense reconstructions then project to each camera image to obtain 2D bounding boxes. (Reconstruction tool from~\cite{Furu:2010:PMVS}.)}
      \label{fig:dense_recon}
   \end{figure}

\section{Data Collection}

        
          
	Our dataset covers a variety of scenes from office buildings and homes, often capturing more than one room. For example a kitchen, living room, and dining room may all be present in one scene. We capture a total of 9 unique scenes, but have a total of 17 scans since some scenes are scanned twice. Each scene has from 696-2,412 images, for a total of 20,916 images and 54,247 bounding boxes. We use the Kinect v2 sensor and code from ~\cite{iai_kinect2} for collection. 

	As stated, we aim to be able to simulate robotic motion through each scene with our scans. At first it may seem the best way to do this is to capture video as the camera moves around the scene. However, in order to get more than one view at any given point the camera must be rotated at that point. Itois not possible to visit the infinite number of points in each scene, so a discrete set of points must be chosen. In a video, even if a consistent frame rate and rotation speed are maintained, there will be images in between the points of rotation that still represent only a single view of a position in the scene. This is unnatural for movement. Imagine a robot arriving at a location and being unable to turn in place.  
    
    We choose to have the camera visit a set of discrete points throughout the scene in order to provide some consistency among the images and camera positions. A video could still be collected at each point of rotation, but this would increase the dataset size unnecessarily. We choose to sample every 30 degrees at each point of rotation, providing substantial overlap between images while keeping the number of images in each scene manageable. 
                       
                       The set of points our robot visits in each scene is essentially a rectangular grid over the scene. We make our points 30 centimeters apart, and justify this in later experiments. Our scenes have between 58-201 points, which allow many choices of how to move. 
                       

               	Two scans of a scene will have different placements of objects. Only objects that would be naturally moved in daily life are relocated. For example chairs, books, and BigBIRD objects may be moved, but sofas and refrigerators will stay put. There are two advantages to scanning each scene twice. First, we are able to get more data from each scene, which is important given the limited availability of scenes. 
Second, we can test a system that learns about objects and a scene from an initial scan, and then is tested on the same scene with moved or new objects, e.g. ~\cite{RobotInARoom}.                

            \begin{figure}[thpb]
      \centering
     \includegraphics[width=1.0\linewidth]{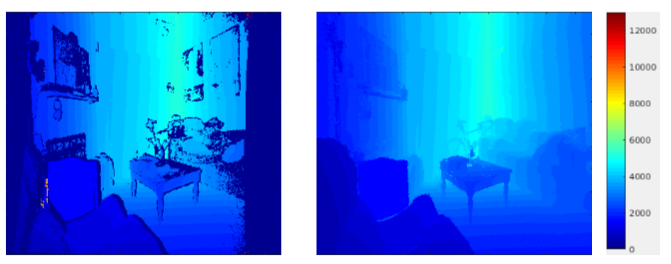}
      \caption{A comparison between an initial depth image(left) and the improved depth image(right). The improved depth images allow us to better handle occlusion when projecting point cloud labels from the dense reconstruction to bounding boxes in the RGB images.}
      \label{fig:depth_improvement}
   \end{figure}

        \subsection{Labels}
        We aim to collect 2D bounding boxes of our 33 common instances across all scenes. In addition, we need to provide movement pointers from each image to allow movement through the scene. We provide pointers for rotation clock-wise and counter clock-wise, as well as translation forward, backward, left, and right.
        
        For each scan of each scene, we create a sparse reconstruction of the scene using the RGB structure from motion tool COLMAP from Sch\"{o}nberger et al \cite{schoenberger2016sfm,schoenberger2016mvs}. From the reconstruction we get the camera position and orientation for each image. We don't use depth information for the reconstruction because our sampling is so dense that we are rarely testing the limits of the RGB system. See Figure \ref{fig:cam_poses} for example reconstructed camera positions.  
              
                             Using the camera positions and orientations we are able to calculate the movement pointers that allow navigation through each scene using natural robotic movements. 

        To label every object instance in each scan, we feed the output of COLMAP into the dense reconstruction system CMVS\//PMVS \cite{Furu:2010:PMVS,Furu:2010}. This gives us a denser point cloud of the scene that makes it easy for humans to recognize objects. We then extract the point cloud of each instance from this dense reconstruction, and are able to get 2D bounding boxes in every image by projecting the point clouds for each object into each image. See Figure~\ref{fig:dense_recon}. 
        Given that most of our scans include multiple rooms and lots of clutter, we must account for occlusion or the point clouds will project through walls and occluding objects and give low quality 2D bounding boxes. We are able use the Kinect depth maps with the reconstructed point clouds and camera poses to account for some occlusion, but not all. Some occlusion is missed by the raw depth maps because they are sometimes noisy, giving wrong or no values for reflective\/shiny surfaces, and are not at the same resolution as the RGB images. 

        To improve a given depth map $D$, we build a dense reconstruction by back projecting the depth maps of cameras that see similar areas of the scene. This solves the difference in resolution problem, as the other depth maps cover the areas missed by $D$. We are also able to fill in many of the missing or wrong values on specular surfaces by taking advantage of the fact that these values are either zero, or much greater than the true depth. Each depth image has a slightly different view of the specular surface, and so has various correct and incorrect values on that surface. By projecting the point clouds of many depth images into $D$ and keeping the smallest value for each pixel, we are able to remove most of the wrong values that are too large, and fill in a lot of the missing values. As a last step we perform some simple interpolation to attempt to fill in any holes of missing values that are left. See Figure \ref{fig:depth_improvement} for a comparison of original to improved depth maps. 
        
       Though the improved depths are much better they are still not perfect. There is also noise in the dense reconstruction and noise in the labeled point clouds. Knowing this, we inspect every bounding box ourselves to make sure it contains the correct object, and is not of poor quality (too large or small for the object). We have labeled our  scans for BigBIRD objects, yielding an average of over 3000 2D bounding boxes per scan. We provide some measure of difficulty for each bounding box based on its size, leaving adding a measure of occlusion for future work. For our experiments we only consider boxes with a size of at least $50x30$ pixels.

     \begin{figure}[thpb]
      \centering
     \includegraphics[width=1.0\linewidth]{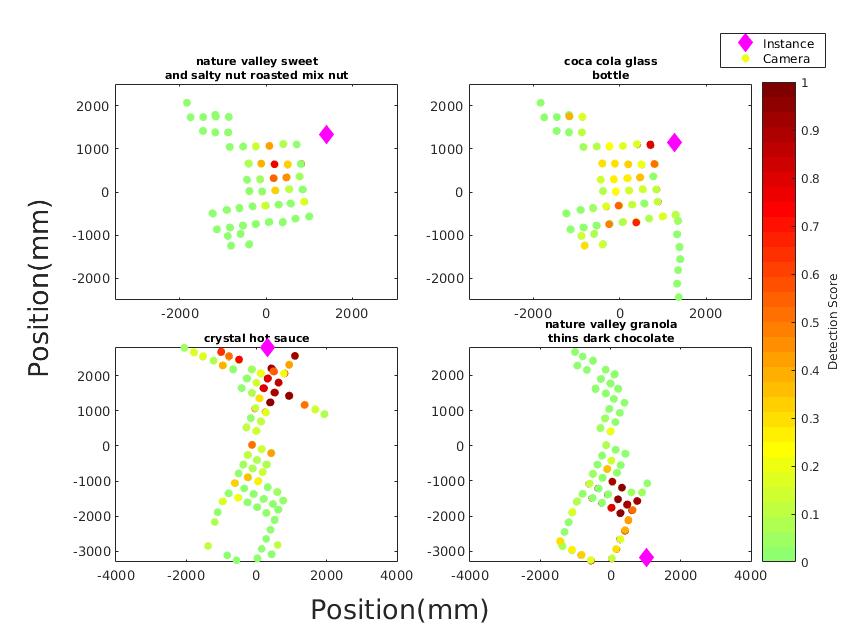}
      \caption{Detection scores for four different instances in various scenes. Dots are camera position, color indicates score. Only cameras that see the instance (purple diamond) are shown. Notice certain viewpoints consistently yield higher scores. It would be advantageous for a robot to move from green views to red ones. }
      \label{fig:detection_changes}
   \end{figure}
    \section{Experiments}

  We aim to show four things: a baseline for instance detection on our data, why it is important to design systems specifically for robot motion, how our dataset can be used to simulate motion, and a system demonstrating an active vision task on our dataset. 
  
  \subsection{Instance Detection}\label{sec:detection}
  We use a state-of-the-art class level object detector as a baseline for instance detection on our dataset. We choose the Single Shot Detection (SSD) network from \cite{liu15ssd} because it offers both real time detection performance (72 FPS) while maintaining a high-level of accuracy. This is exciting for robotics applications for which real time performance is crucial. 
The SSD network consists of a base network, in our case VGG~\cite{simonyan2014very}, with additional feature maps added on top of the base network through a series of 1x1 and 3x3 convolutions.
  
  
  We separate our dataset into three training and testing splits. Each split consists of eleven scans from seven scenes as training and three scans from two  scenes for testing. Since small objects present a particularly difficult challenge for our detector, we first only consider boxes of size at least $100x75$ pixels for training and testing. We then include all boxes of size at least $50x30$, adding more training data but also a more difficult test scenario. 
  
  
  We use 500x500 images for training SSD. We train the network using an initial learning rate of 0.001 and train the network for 20,000 iterations with a stepsize of 6,000. We choose to use the same hyperparameter settings across all splits of the data. The  Mean Average Precision results for each split are shown in Table \ref{tbl:detections}. From this table we can see that the network's performance can vary depending upon the training and testing split used. In the next section we explore how the detection performance is affected by numerous factors in our dataset.

  \subsection{Qualitative Results}\label{sec:det_variation}

\begin{figure*}[thpb]
      \centering
     \includegraphics[width=1.0\linewidth]{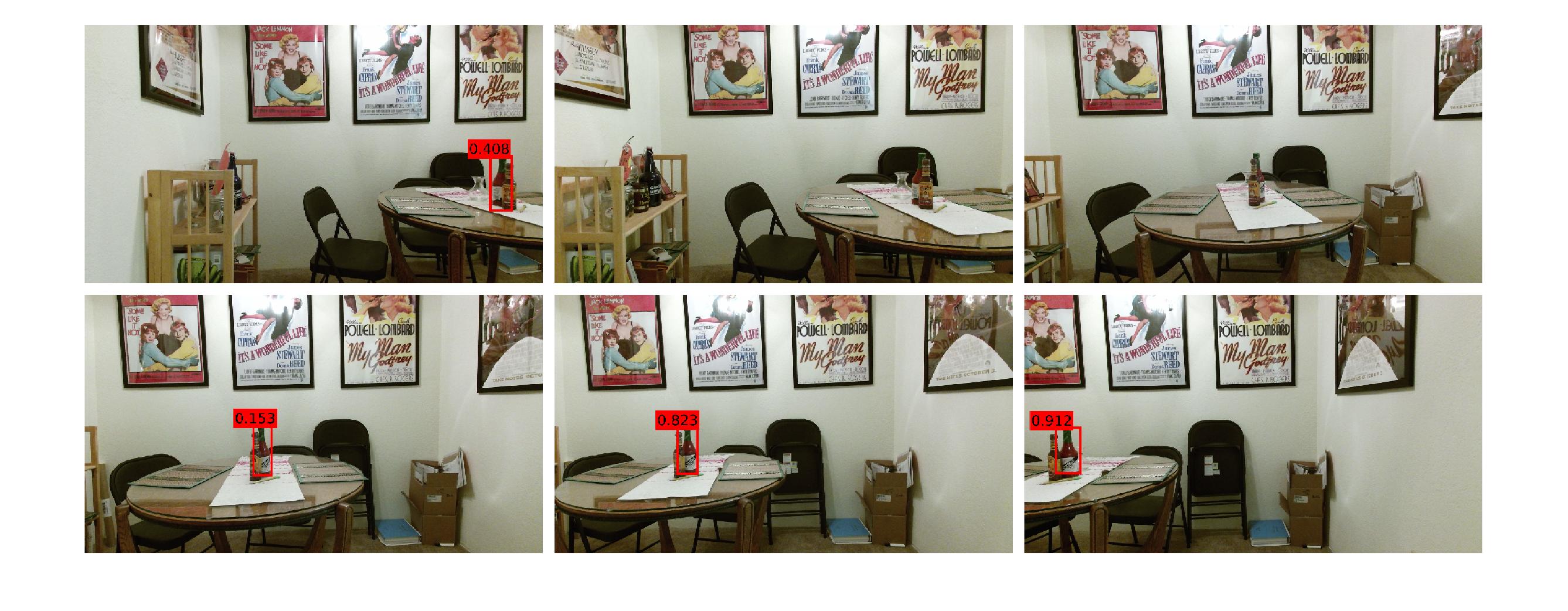}
      \caption{Example of how movement affects detection output for a single instance. The proposed box with highest score $>.1$ for the crystal hot sauce bottle instance is shown in each image. Object instance and scene correspond to the bottom left plot in Figure~\ref{fig:detection_changes}}
      \label{fig:detection_ex}
   \end{figure*}

    As our data has a wide variety of views of each object, varying pose and scale, we wanted to see how the detector fared with respect to different views. Figure \ref{fig:detection_changes} shows how detection score changed when camera position changed relative to an object instance. We can see  there is a clear pattern showing the detector is more reliable in some camera positions than in others.  Figure \ref{fig:detection_ex} shows how occlusion and object pose can greatly impact the detector even though there are training examples for both cases. We observed similar performance for many objects in all of our test scenes. This behavior motivates an active system that can move from a position with poor detection outputs to one with improved performance.

\begin{table}
	\begin{center}
    \begin{tabular}{|l|c|c|c|}
  \hline
  Instance & Split 1 & Split 2 & Split 3  \\
  \hline\hline
  $Boxes > 100x75$ & .39 & .55& .53 \\
  \hline
  $Boxes > 50x30$ & .26 & .41& .42 \\
  \hline
	\end{tabular}
	\caption{MAP detection results. Since small boxes are challenging for detection systems to reproduce, we train/test our detector first using only boxes of size at least $100x75$, and then re-train/test on all boxes at least $50x30$.}
    \label{tbl:detections}
	\end{center}
\end{table}



                                 \begin{figure}[thpb]
      \centering
     \includegraphics[width=1.0\linewidth]{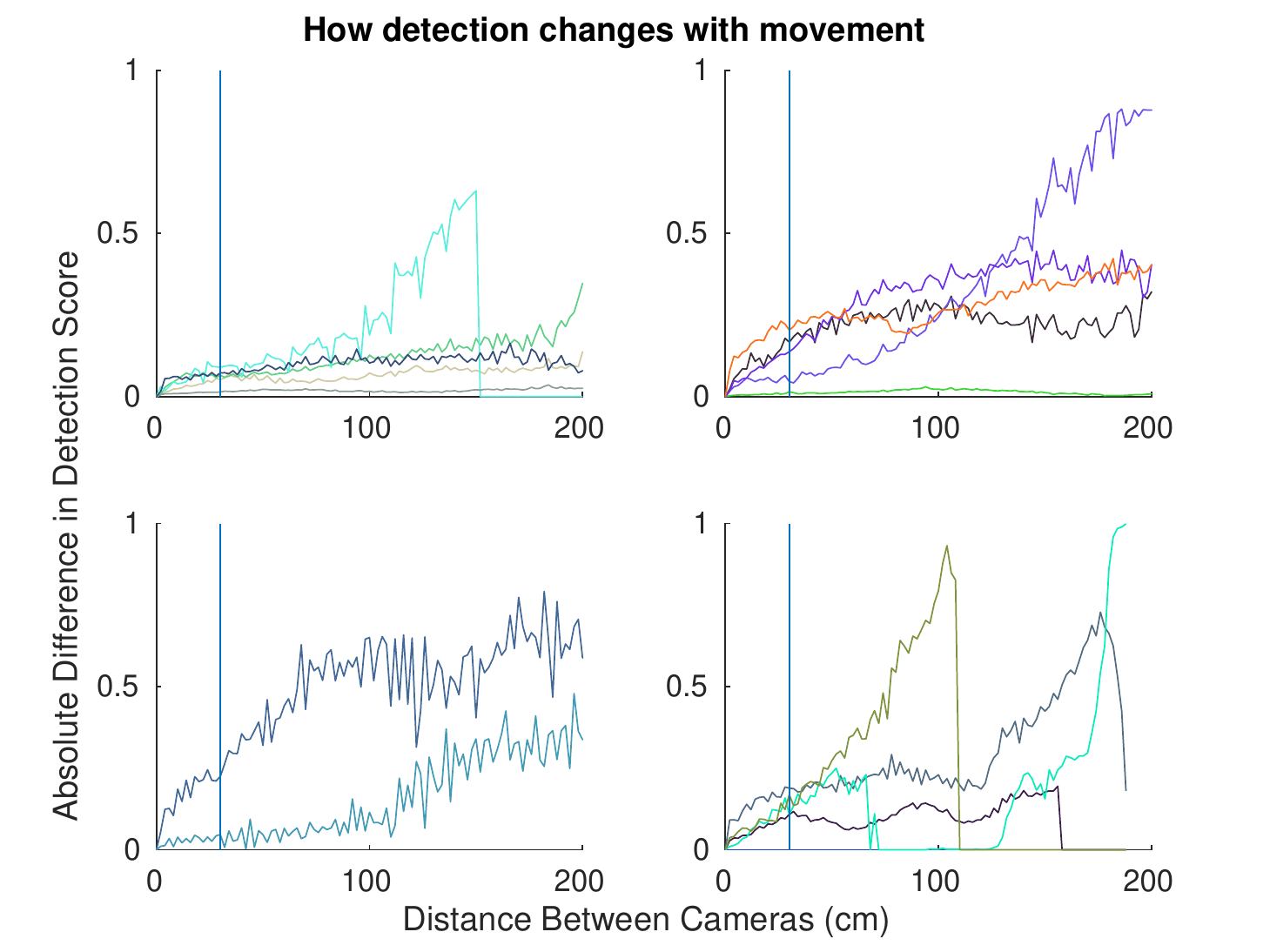}
      \caption{How sensitive our detection system is to change in camera position. As the distance between two images of an instance increases(x-axis), the change in detection score(y-axis) tends to increase. Each line represents one instance. The vertical blue line shows our chosen sampling resolution of 30cm.}
      \label{fig:density}
   \end{figure}

  \subsection{Ability to Simulate Motion}\label{sec:density}

   %
    There are many parts of a robotic system that may be impacted by movement, but we are focused on the vision system, in particular object recognition. To find an appropriate sampling resolution for object recognition, we see how a vision system's output changes as a function of camera movement. We need to find a sampling resolution that can simulate motion but is also practical for data collection purposes. 
    
We first drive our robot around some scenes, capturing video as if the robot is naturally moving through the environment. We then label all BigBIRD instances in the videos, and run our instance detector on each image. For each video, we calculate the difference in detection score for each instance in all pairs of images.  For example, we take the fourth and tenth frame and plot the difference in score for an instance against the distance the camera moved between frames. 
 We plot the results from four videos in Figure \ref{fig:density}.

 For all instances that were detected in at least one image (score greater than 0), even the smallest movement of the camera results in some change in detection score. As the distance between cameras increases, there is a greater change in detection score. We considered the trade-off of having lower variation in our vision system against practicality of data collection.  The vertical blue line in each plot in Figure \ref{fig:density} shows our chosen resolution, 30 cm. We found that for most instances, the change in score at 30cm is not much different than the changes at smaller resolutions like 10 or 20cm.

    \subsection{Active Vision}\label{sec:nbv}
   
         \begin{figure*}[thpb]
      \centering
     \includegraphics[width=1.0\linewidth]{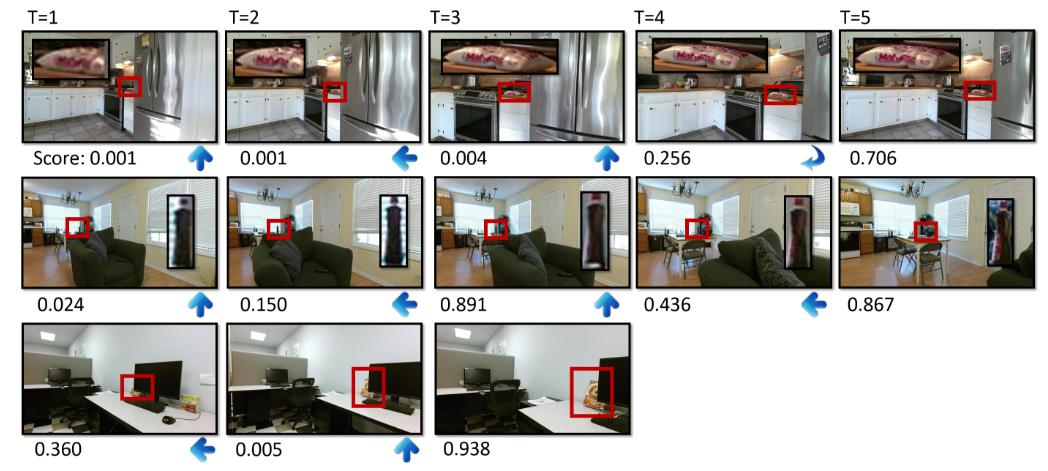}
      \caption{Example paths taken by our active vision system. The arrow indicates the action chosen by the action network. }
      \label{fig:examples}
   \end{figure*}
    
    In this section we propose a baseline for an active instance classification task on our dataset. 
     We envision a scenario where a robotic system is given an area of interest, and the system must classify the object instance at that location. We assume that given an initial area, localizing the same area in subsequent images is straight forward. Based on these assumptions, we propose the following problem setting. As input our agent receives an initial image with a bounding box for the target object. The agent can then choose an action at each timestep and will receive a new image and bounding box corresponding with the action. The goal is for the agent to learn an action policy which will increase the accuracy of the instance classifier.
    
    A straightforward way of training an active vision system for object recognition would be to train the system to acquire new views of an object when there is occlusion. However, it is not easy to label and quantify the level of occlusion of a target object. Furthermore, even if these labels were readily available our intuition about which views are difficult for a classifier would not necessarily be correct. For example, a classifier may be able to easily recognize some heavily occluded objects by only looking at some small discriminative part of the object. In addition, our  dataset contains numerous factors which make the classification task difficult in addition to occlusions, such as varying object scale and lighting conditions. Therefore, we choose to use classification score as the training signal for our active vision system. A new view of an instance can increase both the confidence and accuracy of our classifier. This leads our model to learn a policy which  attempts to move the agent to views that improve recognition performance.
    
\begin{figure}[thpb]
   \centering
   \includegraphics[width=1.0\linewidth]{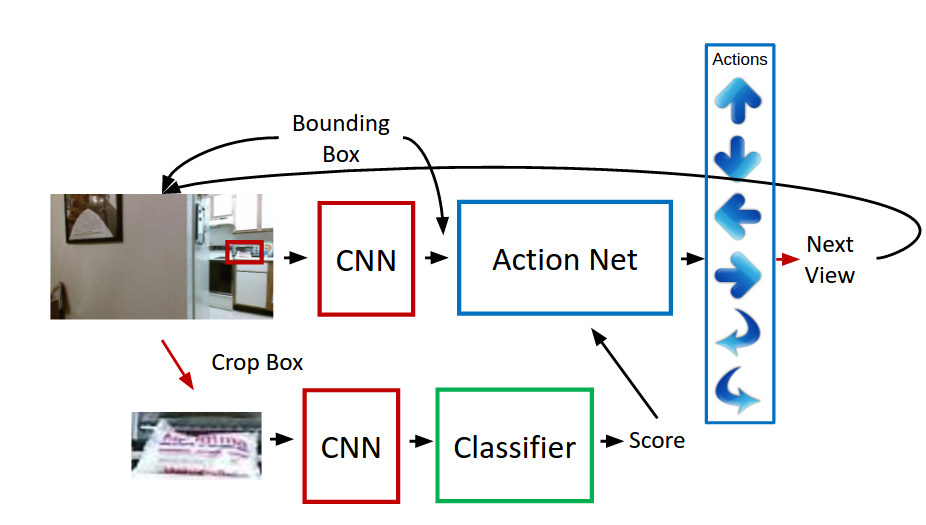}
   \caption{Overall architecture of our active recognition system. It consists of three components. A CNN for extracting image features from the entire image given the current view, an instance classifier for classifying the cropped object, and an action network for selecting the next action in order to improve classification.}
   \label{fig:action_net}
   \end{figure}

    As a feature extractor, we used the first 9 convolutional layers of ResNet-18 models ~\cite{He_2016}, which recently showed compelling results on the 1000 way imagenet classification task. We used pre-trained models written in the torch framework ~\cite{torch}. The weights for the network are fixed for all experiments although our overall system is end-to-end trainable. The instance classifier and action network share the feature extractor. See Figure~\ref{fig:action_net}.
    
    We first train an instance classifier for BigBIRD~\cite{BigBIRD} instances, which appear in our dataset. One natural choice might be to train the classifier and action network simultaneously on our dataset. However, deep neural networks can easily achieve almost 100\% classification accuracy on our training dataset. This type of over-fitting would prevent our action network from learning a meaningful policy, and does not perform well on the test set. 
    
     Thus, we use images from the BigBIRD~\cite{BigBIRD} dataset for training our instance classifier. Even though the BigBIRD dataset provides many viewpoints of an instance, it can't be directly used for training since it consists of objects against a plain white background. We instead use the provided object masks to crop the object and overlay it on a random background sampled from SUN397 dataset\cite{Xiao_2010,RENDER-4-CNN}. In order to prevent our network from overfitting, we aggressively applied various data augmentations. These included randomly cropping part of the image, performing color jittering, and sampling different lightening. Additionally, since our dataset consists of many small object instances, we randomly scaled the object by a factor ranging from 0.02 to 1.

    Our baseline action network is inspired by a recent active vision approach~\cite{Jayaraman_2016,Pairwise_active}. We use the REINFORCE algorithm to train a network to predict an action at each time step. At each time step our action network receives as input an image and a bounding box for the current position. Our network then outputs a score for each action: forward, backward, left, right, clockwise rotation, and counter-clockwise rotation. We fix the maximum number of timesteps during training to be $T=5$ steps or until the classifier achieves more than 0.9 confidence score. If the instance classifier correctly classifies the instance at the final timestep or reaches a 0.9 score at any timestep, we consider the actions taken by the action network as correct. We then give the network a positive reward signal to adjust the weights of the action network to encourage the chosen moves.

    More formally, we want to maximize the expected reward with respect to the policy distribution represented by our action network. 
\begin{equation}
J(\theta) = \mathbb{E}_{p(a_{1:T}|\phi(I_{1:T}),bb_{1:T};\theta)}[R]
\end{equation}
Where $\phi(I_{1:T})$ are the CNN features for the images, $bb_{1:T}$ are the bounding boxes of target objects. If the classification is correct $R$ is the score of the classifier, otherwise $R=0$.
For simplicity, we assumed the policy distributions to be independent at each timestep, $p(a_{1:T}|\phi(I_{1:T}),bb_{1:T};\theta) = \prod_t^T p(a_t|\phi(I_{t-1}),bb_{t-1};\theta)$. 
In order to compute gradients with respect to the parameters of our action network, we use the REINFORCE algorithm, which is sample approximation to the gradient introduced by \cite{Williams_1992} and recently popularized by \cite{Minh_2014}. 
\begin{equation}
\nabla_{\theta}J \approx \frac{1}{M}\sum_{i=1}^{M} \sum_{t=1}^{T}\nabla_{\theta}\log p(a_t^i|\phi(I_{t-1}^i),bb_{t-1}^i;\theta)R^i
\end{equation}
    
    We evaluate our action network by comparing the accuracy of our classifier at different timesteps. The action network is used to choose an action at each image location at each time step, moving to a new image location for the next timestep. We consider how the classification accuracy changes as the maximum timestep, $T$, increases.
    
\begin{figure}[thpb]
      \centering
     \includegraphics[width=1.0\linewidth]{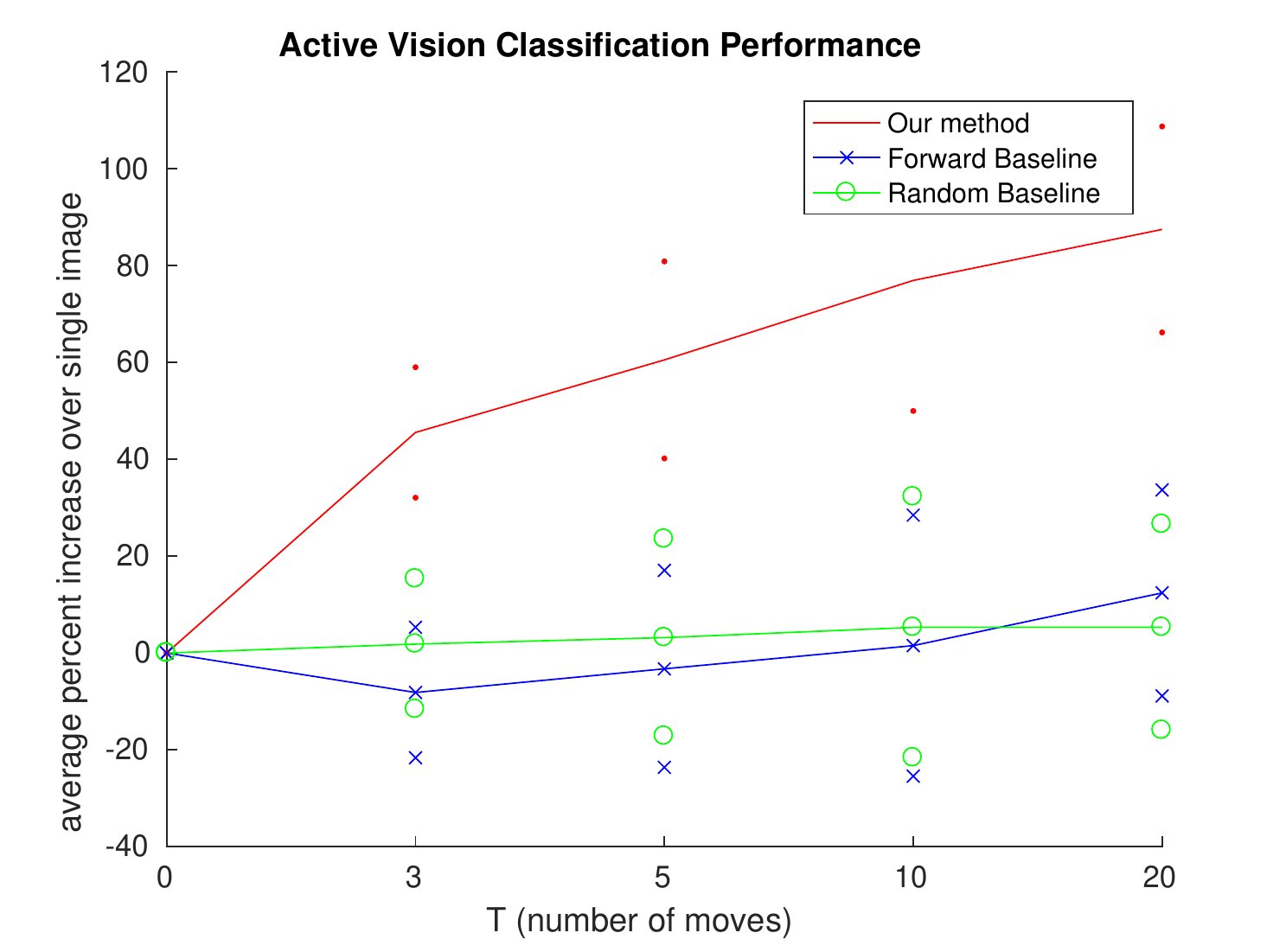}
      \caption{The relative improvement in classification accuracy for different active vision policies. As the system makes more virtual moves through the scene(T increases), our method is able to move to a position that increases classification performance. Making random moves, or just moving forward, does not improve performance much. }
      \label{fig:active_plot2}
   \end{figure}

    Since many of the instances in our dataset are small and far away in the image a natural baseline policy is one that always chooses the move forward action. We additionally compare against a policy of choosing a random action. Figure \ref{fig:active_plot2} shows how our system is able to greatly improve classification accuracy by moving to new image locations. We are also able to outperform the two obvious baselines. Figure~\ref{fig:examples} shows some qualitative examples of our system moving through a scene.
    
    One potential improvement to our active classification model is a method for aggregating the views at each time step in order to choose the next action and perform multi-view classification\cite{su15mvcnn,Pairwise_active}. We also would like to explore the recurrent models that could consider history of actions taken. Additionally, the active vision task difficulty can be further increased by not providing the bounding box. This would require a policy that considers several hypothesis of both the location and class of the object. We expect that our dataset will provide a challenging test bed for further active vision research.

\begin{table}
\begin{center}
\begin{tabular}{|l||c|c|c|c|c|}
\hline
Number of Moves & 0 & 3 & 5 & 10 & 20\\
\hline
\hline
Method &\multicolumn{5}{|c|}{Split 1}\\
\hline
\hline
Ours & .30 & .43 & .45 & .49 & .51 \\
\hline
Random & .30 & .26 & .28 & .28 & .33 \\
\hline
Forward & .30 & .29 & .29 & .29 & .29 \\
\hline
&\multicolumn{5}{|c|}{Split 2}\\
\hline
\hline
Ours & .25 & .40 & .46 & .52 & .53 \\
\hline
Random & .25 & .24 & .26 & .29 & .33 \\
\hline
Forward & .25 & .29 & .30 & .31 & .31 \\
\hline
&\multicolumn{5}{|c|}{Split 3}\\
\hline
\hline
Ours & .42 & .56 & .62 & .67 & .73 \\
\hline
Random & .42 & .38 & .40 & .42 & .46 \\
\hline
Forward & .42 & .39 & .39 & .40 & .40 \\
\hline
\hline
\end{tabular}
\end{center}
\caption{Active vision results for different splits. Columns represent number of moves. Numbers are accuracy of the classifier, averaged across all instances in all test scenes. The goal of our system is to move in the scene to increase classification accuracy for a particular instance. }
\label{tbl:active}
\end{table}


\section{Conclusions}
We introduce a new labeled dataset for developing and benchmarking 
object recognition methods in challenging 
indoor environments and active vision strategies for these tasks. 
We establish a baseline for object instance detection and show
that the data is suitable for training a modern deep-learning-based 
system for next best view selection, using reinforcement learning, something that usually requires using a robot in the loop or synthetic computer graphics models. Using our densely sampled RGB-D imagery
allows systems to see and be evaluated on real-world visual perception
challenges which include large variations in scale and viewpoint 
as well as real imaging conditions that may not be present in CG.  
We validate experimentally that current state-of-the-art detection 
systems benefit from active vision on this real-world data.  
The dataset and toolbox for processing are now public.

\bibliographystyle{IEEEtran}
\bibliography{rohit_bib}
\end{document}